\documentclass[letterpaper]{article}

\usepackage{ifpdf}

\ifpdf
  \usepackage[preprint]{aaai2027}
\fi

\usepackage[hyphens]{url}
\usepackage{graphicx}
\usepackage{natbib}
\usepackage{caption}
\usepackage{booktabs}
\usepackage{amsmath}
\usepackage{amssymb}
\usepackage{algorithm}
\usepackage{algpseudocode}
\usepackage{textcomp}

\ifpdf
\fi

\title{CURE: Local Uncertainty Repair for Block-Parallel Speculative Decoding}

\ifpdf
  \author{
    Aofan Liu\equalcontrib,
    Jingxiang Meng\equalcontrib,
    Fangxin Liu\corresponding,
    Yongbiao Chen\corresponding
  }
  \affiliations{}
\else
  \author{
    Aofan Liu\textsuperscript{*},
    Jingxiang Meng\textsuperscript{*},
    Fangxin Liu\textsuperscript{\textdagger},
    Yongbiao Chen\textsuperscript{\textdagger}
  }
  \date{}
\fi

\begin{document}

\maketitle

\ifpdf\else
  \begingroup
  \centering
  \footnotesize
  \textsuperscript{*} These authors contributed equally.\quad
  \textsuperscript{\textdagger} Corresponding authors.
  \par
  \endgroup
\fi

\begin{abstract}
Speculative decoding mitigates the latency of sequential generation in autoregressive Large Language Models (LLMs) by interleaving draft generation with target verification. However, existing parallel drafting backends often suffer from rapid accuracy degradation over long horizons, leading to high rejection rates during verification and suboptimal wall-clock speedups. We observe that drafting errors are not uniformly distributed but typically stem from localized high-uncertainty tokens that destabilize downstream generation trajectories. Motivated by this token error pattern, we propose CURE, a budget-aware dynamic repair tree designed to repair errors at uncertainty focal points without incurring prohibitive tree-verification overheads. Specifically, our method uses predictive confidence margins to dynamically locate candidate error tokens within a block-parallel draft, expands bounded repair paths only at these fragile nodes, and employs a novel repair resynchronization mechanism to realign draft states post-verification. Evaluations on code-generation benchmarks (HumanEval, MBPP, and LiveCodeBench-lite) and mathematical reasoning benchmark (GSM8K) demonstrate that CURE increases the average accepted length by 4.2--7.5\% over parallel baselines without repair, translating to an end-to-end speedup of 2.66--3.49\texttimes{} over target-only decoding. Furthermore, we provide a plug-and-play repair module compatible with standard parallel drafting frameworks. We also characterize the trade-off between draft compute and verification efficiency.
\end{abstract}

\section{Introduction}

Autoregressive inference in Large Language Models (LLMs) suffers from high latency due to sequential, token-by-token generation. Speculative decoding mitigates this memory-bandwidth bottleneck by leveraging a lightweight draft model to generate candidate tokens, which are subsequently verified in parallel by a target LLM~\citep{leviathan2023fast,chen2023accelerating,liu2024speculative}. Recent advancements expand this paradigm either by constructing candidate trees or by learning specialized draft representations~\citep{cai2024medusa,li2024eagle}. Among these, \textit{block-parallel drafting} has emerged as an appealing paradigm, generating multi-token blocks in a single forward pass without sequential drafting overhead.

Despite its computational efficiency, block-parallel drafting exhibits an inherent structural fragility: \textbf{the error propagation cascade}. Because block-parallel models generate tokens independently or in coarse blocks, a single localized error at a low-confidence position causes conventional prefix verification to reject the entire subsequent draft sequence, even if the trailing tokens are semantically correct. While unconstrained tree-based speculative decoding can recover alternative trajectories, applying dense tree expansion uniformly across all blocks incurs severe target-verification overhead and memory footprint, wasting compute on predictable spans that require no repair.

In this work, we argue that candidate tree expansion should be \textbf{frugal and spatially localized}. Taking code generation as a representative domain, source code exhibits strong global structural predictability~\citep{hindle2012naturalness}, yet local variations (such as identifiers, API calls, and syntax operators) introduce localized uncertainty~\citep{casalnuovo2020predictable}. We observe that draft errors are not uniformly distributed; rather, they originate from a few \textit{Uncertainty Focal Points (UFPs)}. Identifying and repairing these specific focal points can prevent downstream cascade rejections without triggering expensive full-tree construction.

To operationalize this intuition, we propose \textbf{CURE} (Local Uncertainty Repair), a budget-aware dynamic repair tree built atop block-parallel draft backends. Prior to candidate verification, our framework evaluates the top-1/top-2 log-probability margins of the block-parallel draft to pinpoint local UFPs. It then dynamically allocates a bounded verification budget, spawning alternative candidate branches strictly at these fragile positions. To ensure seamless execution, we introduce a \textit{cache-resynchronization mechanism} that realigns the draft model's internal state with the target-verified path upon token acceptance, eliminating the need for draft model retraining.

We conduct extensive evaluations on primary code generation benchmarks (\textit{HumanEval, MBPP, LiveCodeBench-lite}) and out-of-domain reasoning (\textit{GSM8K}). Experimental results show that \textbf{CURE} consistently increases the accepted tokens per step by 4.2\% to 7.5\% over single-path parallel backends, achieving a $2.66\times$ to $3.49\times$ end-to-end speedup over target-only autoregressive decoding. Crucially, our empirical analysis reveals a fine-grained \textbf{acceptance--compute Pareto trade-off}: while local repair expands the accepted context length, the additional tree-verification compute presents a tunable knob between arithmetic intensity and memory-bound latency, offering valuable insights for hardware-aware speculative decoding design.

Our main contributions are summarized as follows:
\begin{itemize}
\item \textbf{Localized Repair Formulation:} We formulate draft candidate expansion as a constrained budget-allocation problem, introducing a confidence-margin gating mechanism that confines alternative branching strictly to localized Uncertainty Focal Points (UFPs).
\item \textbf{Training-Free \& Cache-Consistent Architecture:} We design a plug-and-play dynamic repair tree with a novel cache-resynchronization procedure. It enables training-free integration with block-parallel backends while maintaining KV-cache consistency across draft-target state transitions.
\item \textbf{Pareto Analysis of Acceptance vs. Cost:} We analyze the acceptance-latency trade-off across multiple code and reasoning benchmarks, establishing quantitative guidelines on when local branching yields net wall-clock acceleration versus when it incurs verification saturation.
\end{itemize}

\section{Related Work}

Speculative decoding separates generation into proposal and verification. Draft-then-verify generation was first developed for sequence-to-sequence decoding and later formalized as lossless sampling from an autoregressive target model~\citep{xia2023specdec,leviathan2023fast,chen2023accelerating}. A smaller model proposes several tokens, and one target-model forward pass accepts a prefix and corrects the first rejection. Although this procedure preserves the target distribution, its wall-clock speedup depends jointly on draft latency, verification latency, proposal length, and accepted tokens~\citep{liu2024speculative,yan2024decoding}.

Draft-model research therefore balances proposal cost against target agreement. DistillSpec uses on-policy knowledge distillation and task-dependent divergence objectives, Online Speculative Decoding adapts drafters to observed queries, and HASS aligns representations and token-level objectives~\citep{zhou2024distillspec,liu2024online,zhang2025hass}. Other methods alter the draft architecture. Medusa adds multiple prediction heads to the target, Hydra conditions later heads on earlier draft tokens, and EAGLE performs autoregression in the target feature space~\citep{cai2024medusa,ankner2024hydra,li2024eagle}. EAGLE-3 replaces feature prediction with direct token prediction and fuses multiple target layers~\citep{li2025eagle3}. Self-speculative methods skip target layers or exit early, while REST retrieves candidate continuations from prior text~\citep{zhang2024draft,elhoushi2024layerskip,he2024rest}. These approaches improve acceptance or reduce draft cost through training, adaptation, architectural changes, or retrieval. In contrast, \textbf{CURE} leaves the pretrained block-parallel drafter fixed and alters how uncertain draft positions are repaired at inference time without requiring architectural changes or retraining.

\begin{figure*}[t]
\centering
\includegraphics[width=\textwidth]{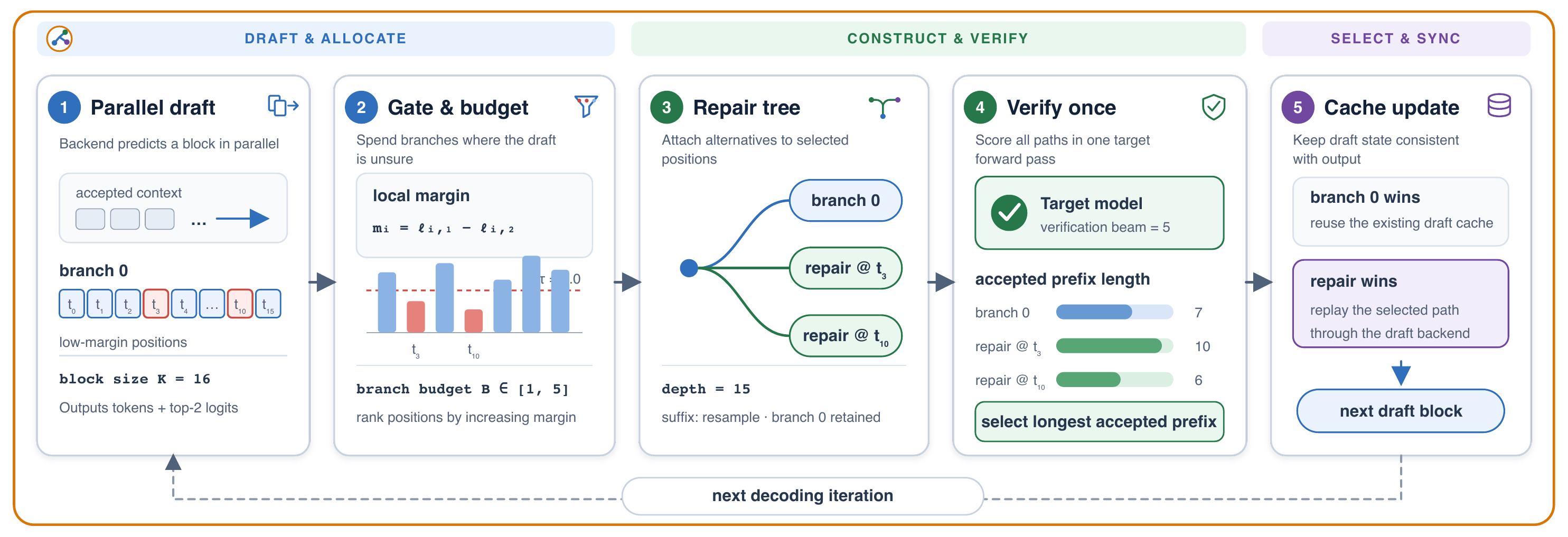}
\caption{Overview of one decoding iteration of \textbf{CURE}: a block-parallel draft produces branch 0 alongside per-position uncertainty margins. Low-margin positions receive a bounded repair budget, and the target model verifies branch 0 together with the resulting repair paths in a single pass. The framework selects the longest accepted prefix and resynchronizes the draft cache only when a repair path wins.}
\label{fig:method}
\end{figure*}

Parallel and non-autoregressive drafting instead reduces the number of drafter calls. Blockwise parallel decoding predicts several future positions simultaneously and accepts a verified prefix~\citep{stern2018blockwise}. Lookahead decoding constructs and verifies Jacobi-style trajectories without a separate draft model, and Lookahead Reasoning extends step-level speculation to long reasoning traces~\citep{fu2024lookahead,fu2025lookaheadreasoning}. Recent learned drafters replace token-by-token proposals with block-level computation. DFlash uses block diffusion conditioned on target features, whereas DART predicts several masked positions in parallel and prunes the resulting tree with an n-gram constraint~\citep{chen2026dflash,liu2026dart}. Domino adds causal correction to a parallel backbone so that each position receives prefix-dependent information without returning to full autoregressive drafting~\citep{huang2026domino}. SpecBlock takes a hybrid approach, generating blocks of dependent positions and constructing a dynamic tree through repeated block expansion~\citep{shi2026specblock}. These methods expose a fundamental trade-off: independent parallel predictions are computationally cheap but lose path consistency, whereas stronger within-path dependence improves proposal quality at added draft cost. \textbf{CURE} operates after a parallel block is generated, preserving the low-cost path and spending additional verification budget strictly where the block reports local ambiguity.

Adaptive candidate construction provides another way to allocate compute. Fixed proposal lengths and static tree shapes spend identical compute on easy and difficult contexts. SpecDec++ learns when to stop autoregressive drafting from conditional acceptance probabilities, while AdaEAGLE predicts a context-dependent draft length~\citep{huang2024specdecpp,zhang2024adaeagle}. Multi-candidate methods instead alter the verification structure. SpecInfer merges proposals into a token tree, GraphSD shares repeated subsequences in a graph, and SpecTr formulates verification through optimal transport~\citep{miao2023specinfer,gong2024graph,sun2023spectr}. SEQUOIA optimizes tree construction under serving constraints, while SLiM removes low-utility hypotheses before verification~\citep{chen2024sequoia,lin2024slim}. Other adaptive trees use draft confidence to decide structure growth: DySpec greedily expands nodes according to draft probabilities, OPT-Tree adapts its topology to the draft distribution, and EAGLE-2 constructs dynamic trees from confidence estimates~\citep{xiong2024dyspec,wang2025opttree,li2024eagle2}. While these methods improve candidate coverage, larger structures increase candidate-construction, attention, and target-verification costs. \textbf{CURE} operates at a distinct boundary: it preserves the block-parallel prediction as branch 0, allocates alternatives exclusively to locally uncertain positions, and replays a winning repair path to maintain draft-cache continuity. It is therefore a selective post-draft repair mechanism under a bounded verification budget rather than a new draft architecture or a general tree-search procedure.

Code generation provides a useful workload for evaluating local repair. Source code contains repeated lexical and syntactic patterns, where local expression choices are often highly predictable~\citep{hindle2012naturalness,casalnuovo2020predictable}. This regularity can yield long speculative matches, yet identifiers, constants, APIs, and formatting create localized ambiguity that may prematurely terminate an otherwise valid prefix. Work on Verilog generation further shows that syntax-aware token boundaries materially impact speculative-decoding performance~\citep{xu2025verilog}. However, token agreement does not equal functional correctness: two programs may differ textually while passing the same tests, and an identical prefix does not guarantee a correct completion. We therefore evaluate whether local repair recovers accepted tokens and improves latency while reporting execution correctness independently from decoding metrics.

\section{Methodology}

\subsection{Problem Formulation}

Let $x_{<t}$ denote the generated context sequence at decoding step $t$. A block-parallel draft model $q$ generates a length-$K$ candidate sequence $\mathbf{y}_{t:t+K-1} = (y_t, y_{t+1}, \dots, y_{t+K-1})$ alongside its corresponding logit distributions in a single forward pass. A target autoregressive model $p$ subsequently verifies candidate trajectories and accepts a prefix. In standard single-path speculative decoding, candidate generation yields a single path $\pi_0$. When an error occurs at position $k$, standard prefix verification rejects all trailing tokens $y_{j}$ ($j > k$), leading to severe truncation of the accepted length.

To recover valid suffixes beyond early rejection points without full autoregressive drafting, we formulate local repair as a \textit{constrained candidate-tree optimization problem}. Let $\Pi = \{\pi_0, \pi_1, \dots, \pi_R\}$ represent a tree of candidate execution paths rooted at $x_{<t}$, where $\pi_0$ denotes the default block-parallel path and $\{\pi_r\}_{r=1}^R$ denote candidate repair branches spawned at identified error points. Let $a(\pi)$ be the accepted token length along path $\pi$, and let $V(\Pi)$ denote the total token budget (i.e., tree node count) presented to the target verifier. The optimal candidate trajectory $\pi^\star$ maximizes the accepted sequence length under a strict verification compute budget $V_{\max}$:
\begin{equation}
    \pi^\star = \arg\max_{\pi \in \Pi} a(\pi) \quad \text{s.t.} \quad V(\Pi) \le V_{\max}.
\end{equation}
From an efficiency perspective, the effective token generation rate scales with $\frac{\mathbb{E}[a(\pi^\star)] + 1}{\mathcal{C}_d + \mathcal{C}_v(\Pi)}$, where $\mathcal{C}_d$ is the constant draft overhead and $\mathcal{C}_v(\Pi)$ is the memory/compute cost of target tree verification. Therefore, candidate tree expansion yields a positive wall-clock trade-off if and only if the gain in accepted tokens outpaces the increased target verification complexity $\mathcal{C}_v(\Pi)$.

\subsection{Method Overview}

To solve the constrained candidate-tree optimization problem under budget $V_{\max}$, \textbf{CURE} introduces an end-to-end, inference-only pipeline that operates directly atop any pretrained block-parallel backend without model retraining. As illustrated in Figure~\ref{fig:method}, each decoding iteration of \textbf{CURE} proceeds through three integrated phases:

\begin{enumerate}
    \item \textbf{Uncertainty Focal Point (UFP) Detection:} The draft model generates a single-pass block proposal $\pi_0$. \textbf{CURE} inspects the per-position top-1/top-2 logit margins to dynamically locate fragile positions (UFPs) where draft confidence drops precipitously.
    \item \textbf{Budget-Aware Dynamic Tree Construction:} Instead of uniform or unconstrained tree expansion, \textbf{CURE} dynamically calculates candidate branch allocations strictly at identified UFPs. It attaches alternative candidate tokens to $\pi_0$, constructing a sparse repair tree $\mathcal{T}$ that satisfies $V(\mathcal{T}) \le V_{\max}$.
    \item \textbf{Single-Pass Verification and Cache Resynchronization:} The target model $p$ verifies all branches in $\mathcal{T}$ concurrently using Tree Attention. If a repair branch $\pi^\star \neq \pi_0$ yields a longer accepted prefix, \textbf{CURE} accepts the repaired path and resynchronizes the draft model's internal KV-cache to realign subsequent drafting states.
\end{enumerate}

\subsection{Local Uncertainty Gating}

For each position $i \in \{0, \dots, K-1\}$ within a generated draft block, let $\ell_{i,1}$ and $\ell_{i,2}$ denote the top-1 and top-2 unnormalized log-probabilities emitted by the draft model $q$, respectively. We define the predictive confidence margin $m_i$ as:
\begin{equation}
    m_i = \ell_{i,1}-\ell_{i,2}.
\end{equation}
A smaller predictive margin $m_i$ indicates that the draft model exhibits lower confidence in distinguishing the leading candidate tokens. Positions with $m_i < \tau_{\text{margin}}$ are identified as Uncertainty Focal Points (UFPs). We select positions in ascending order of $m_i$ to prioritize candidate expansion at the most fragile nodes while capping total tree size to remain within the verifier budget $V_{\max}$.

In our baseline setup, each selected position expands at most $B = 5$ candidate branches, with a maximum tree depth of 15, a target verification beam width of 5, and a repair margin threshold $\tau_{\text{margin}} = 1.0$. To optimize tree efficiency, we further introduce a dynamic branching mechanism that dynamically maps the normalized uncertainty $u_i \in [0, 1]$ to a local branch budget $b_i$:
\begin{equation}
\begin{aligned}
    u_i&=\operatorname{clip}(1-m_i/s,0,1),\\
    b_i&=\operatorname{clip}(\operatorname{round}(b_{\min}+u_i(b_{\max}-b_{\min})),b_{\min},b_{\max}).
\end{aligned}
\end{equation}
where $s$ denotes the margin scale, $b_{\min} = 1$, and $b_{\max} = 5$. This dynamic allocation adjusts only the local tree topology without modifying the underlying draft checkpoint or requiring retraining.

\subsection{Candidate Tree Construction}

For each identified UFP, \textbf{CURE} constructs candidate repair paths by preserving the original prefix while substituting the fragile token with high-scoring alternatives. The remaining suffix tokens are then resampled under the substituted token. Paths sharing a common prefix naturally merge into shared parent nodes within the verification tree, enabling the target verifier to process all candidate paths in a single pass via Tree Attention without independent batch computation. Crucially, the original block-parallel path ($\pi_0$) is anchored as branch 0 and is protected from pruning by the uncertainty gate.

To ensure computational efficiency, candidate construction is strictly bounded by three complementary constraints: the position gate limits where repair branches may originate, the branch budget caps alternatives per selected position, and the verification beam constrains the active tree frontier. In our primary configuration, at most one low-margin position is repaired per block. This localized strategy concentrates the extra verification budget on critical ambiguous nodes rather than enumerating combinatorial paths across the full sequence block.

\subsection{Joint Verification and Cache Resynchronization}

The constructed candidate paths and the default path $\pi_0$ are unified into a sparse verification tree $\mathcal{T}$. A single forward pass of the target model $p$ evaluates all tree positions concurrently via Tree Attention and computes the accepted prefix length $a(\pi)$ for each candidate trajectory $\pi \in \mathcal{T}$. The framework identifies the optimal trajectory that yields the longest accepted prefix:
\begin{equation}
    \pi^\star = \arg\max_{\pi \in \mathcal{T}} a(\pi).
\end{equation}
If branch 0 wins ($\pi^\star = \pi_0$), the current draft internal states are preserved. If a repair branch wins ($\pi^\star \neq \pi_0$), the emitted sequence deviates from the original draft proposal. To maintain state consistency for subsequent drafting steps, \textbf{CURE} replays the selected repair tokens into the draft model's Key-Value (KV) cache, realigning the draft model's internal context representation with the verified output sequence.

\begin{algorithm}[t]
\caption{\textbf{CURE}: Local Uncertainty Repair for Block-Parallel Speculative Decoding}
\label{alg:cure}
\begin{algorithmic}[1]
\Require Target model $p$, block-parallel draft model $q$, context $x_{<t}$, block size $K$, verification budget $V_{\max}$, margin threshold $\tau_{\text{margin}}$.
\Ensure Accepted token sequence $\mathbf{y}_{\text{acc}}$ and updated context $x_{<t+|\mathbf{y}_{\text{acc}}|}$.

\State $\mathbf{y}_{\text{draft}}, \{\ell_{i,1}, \ell_{i,2}\}_{i=0}^{K-1} \gets q(x_{<t})$ \Comment{Generate block-parallel draft proposal}
\State Initialize default path $\pi_0 \gets \mathbf{y}_{\text{draft}}$, candidate tree $\mathcal{T} \gets \{\pi_0\}$
\For{$i = 0$ \textbf{to} $K-1$}
    \State $m_i \gets \ell_{i,1} - \ell_{i,2}$ \Comment{Compute predictive margin}
    \If{$m_i < \tau_{\text{margin}}$ \textbf{and} $V(\mathcal{T}) < V_{\max}$}
        \State Compute branch budget $b_i$ via Eq. (3)
        \State Sample top-$b_i$ alternative tokens $\mathbf{y}_{i}^{\text{alt}}$ from $q$
        \State Spawn repair branches using $\mathbf{y}_{i}^{\text{alt}}$ and attach to tree $\mathcal{T}$
    \EndIf
\EndFor
\State $\mathbf{y}_{\text{accepted}}, \pi^\star \gets \text{TargetTreeVerify}(p, \mathcal{T}, x_{<t})$ \Comment{Single-pass Tree Attention verification}
\If{$\pi^\star \neq \pi_0$}
    \State $\text{ResynchronizeCache}(q, \pi^\star)$ \Comment{Realign draft KV-cache with winning repair path}
\EndIf
\State \Return $\mathbf{y}_{\text{accepted}}$
\end{algorithmic}
\end{algorithm}

\begin{table*}[t]
\centering
\footnotesize
\setlength{\tabcolsep}{3pt}
\begin{tabular}{lrrrrrrrr}
\toprule
& \multicolumn{4}{c}{TPOT $\downarrow$} & \multicolumn{2}{c}{Speedup vs.\ Target AR ($\times$) $\uparrow$} & \multicolumn{2}{c}{Accepted tokens/step $\uparrow$}\\
\cmidrule(lr){2-5}\cmidrule(lr){6-7}\cmidrule(lr){8-9}
Dataset & Target AR & Naive SD & Parallel & Ours & Parallel & Ours & Parallel & Ours\\
\midrule
HumanEval & 30.430 & 42.203 & 6.232 & 8.718 & 4.883 & 3.490 & 7.165 & 7.641\\
MBPP & 30.279 & 49.853 & 7.919 & 11.378 & 3.824 & 2.661 & 5.808 & 6.053\\
LCB-lite & 31.577 & 55.782 & 6.382 & 10.283 & 4.948 & 3.071 & 6.268 & 6.735\\
GSM8K & 107.070 & 155.153 & 16.165 & 29.461 & 6.624 & 3.634 & 8.514 & 11.231\\
\bottomrule
\end{tabular}
\caption{Controlled end-to-end decoding results. TPOT is measured in ms/token. Speedup is computed against target-only autoregressive decoding within the same run protocol.}
\label{tab:main}
\end{table*}

\subsection{Training and Inference Boundary}
While the pretrained block-parallel draft checkpoint requires specialized multi-token training, all components of \textbf{CURE} (including the dynamic repair tree, margin gating, dynamic branch allocation, and KV-cache resynchronization) operate entirely at inference time. These mechanisms introduce zero additional parameters to either the draft or target models and require no retraining or fine-tuning, establishing \textbf{CURE} as a flexible, plug-and-play module for existing parallel drafting backends.

\begin{table*}[t]
\centering
\footnotesize
\setlength{\tabcolsep}{8pt}
\begin{tabular}{llrrrr}
\toprule
& & \multicolumn{4}{c}{Speedup vs.\ Target AR ($\times$) $\uparrow$}\\
\cmidrule(lr){3-6}
Method & Draft structure & HumanEval & MBPP & LCB-lite & GSM8K\\
\midrule
CURE (ours) & Parallel draft + repair & 3.49 & 2.66 & 3.07 & 3.63\\
EAGLE-3 & 16-node tree & 2.17 & 1.93 & 1.80 & 2.21\\
EAGLE-3 & 60-node tree & 2.50 & 2.22 & 2.03 & 2.56\\
DART & 60-node tree & 2.52 & 2.39 & 2.24 & 2.28\\
DFlash & 16-token block & 5.21 & 4.71 & 5.37 & 5.21\\
Domino & 16-token block & 5.89 & 5.53 & 5.27 & 7.92\\
\bottomrule
\end{tabular}
\caption{Reported Qwen3-8B speedups on shared benchmarks. Each value is relative to the target-only baseline in its own evaluation protocol. Related-method values are reported by Domino~\citep{huang2026domino}.}
\label{tab:published_speedups}
\parbox{0.96\linewidth}{\footnotesize CURE uses the controlled measurements in Table~\ref{tab:main}. Related methods use A100 GPUs and a 2048-token generation limit, whereas our runs use a separate implementation and a 256-token limit. The table provides same-target context rather than a controlled ranking.}
\end{table*}

\section{Experiments}

\subsection{Experimental Setup}

\paragraph{Models and Hardware Platform.}
We evaluate our framework using the Qwen3 series models. The target model $p$ is \texttt{Qwen3-8B}, paired with a pretrained \texttt{Qwen3-8B}-based block-parallel draft backend $q$ (block size $K=16$). As an autoregressive draft baseline, we employ \texttt{Qwen3-4B}. All experiments are conducted on a single GPU using bfloat16 precision and FlashAttention-2. Decoding temperature is set to $0$ (greedy decoding) across all benchmark evaluations. All Time-Per-Output-Token (TPOT) metrics are recorded in seconds per token and reported in milliseconds per token (ms/tok) under sequential single-GPU runs to eliminate inter-request variance.

\paragraph{Datasets and Evaluation Metrics.}
We evaluate model performance across three primary code generation benchmarks and one out-of-domain mathematical reasoning stress test:
\begin{itemize}
    \item \textbf{HumanEval}: 164 code-eval prompts for functional python evaluation.
    \item \textbf{MBPP}: 128 sanitized code-eval prompts.
    \item \textbf{LiveCodeBench-lite (LCB-lite)}: 128 problems using the public \texttt{test6.jsonl}.
    \item \textbf{GSM8K}: 128 multi-step mathematical reasoning tasks serving as an out-of-domain test.
\end{itemize}

We evaluate system efficiency and functional correctness using four primary metrics:
\begin{enumerate}
    \item \textbf{Accept/step ($\tau$):} The mean number of tokens accepted per draft/verification step.
    \item \textbf{TPOT (ms/tok):} Mean end-to-end latency per output token (recorded in seconds and reported in milliseconds).
    \item \textbf{Speedup ($\times$):} The TPOT ratio relative to target-only autoregressive decoding ($\text{TPOT}_{\text{Target}} / \text{TPOT}_{\text{Method}}$).
    \item \textbf{Pass@1 (\%):} Code-execution or public-test pass rate, reported separately from token-level exactness.
\end{enumerate}

\begin{figure*}[t]
\centering
\includegraphics[width=\textwidth]{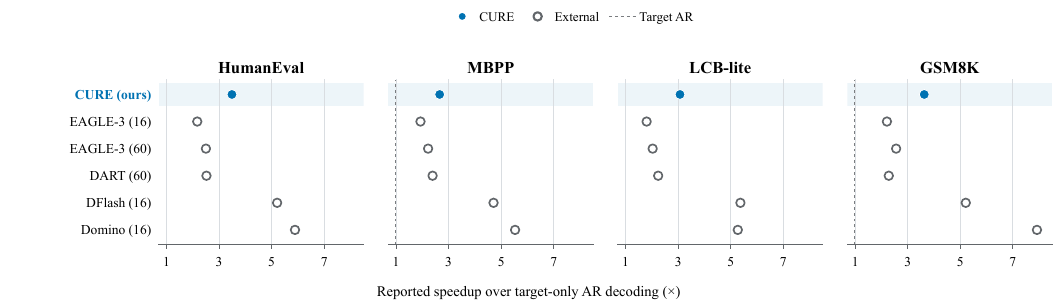}
\caption{Reported Qwen3-8B speedups from Table~\ref{tab:published_speedups}. Filled markers denote CURE; hollow markers denote source-specific external results.}
\label{fig:published_speedups}
\end{figure*}

\paragraph{Baselines.}
We compare \textbf{CURE} against three representative decoding strategies:
\begin{itemize}
    \item \textbf{Target AR:} Standard single-path autoregressive decoding executed directly on Qwen3-8B.
    \item \textbf{Parallel Draft:} The default block-parallel draft backend producing a single candidate path ($K=16$), serving as both the low-cost reference and branch 0 ($\pi_0$) of the repair tree.
    \item \textbf{Naive SD:} Standard speculative decoding drafting each block autoregressively with Qwen3-4B and verifying with Qwen3-8B, measuring the practical cost of token-by-token drafting with a smaller model.
\end{itemize}
\subsection{Main Results}

Table~\ref{tab:main} presents our controlled, same-system evaluation, where \textbf{CURE} is configured with the \texttt{fast\_margin100} variant. To contextualize our results within the broader literature, Table~\ref{tab:published_speedups} separately summarizes reported speedups from existing literature on Qwen3-8B. We utilize the speedup ratio relative to each benchmark's target-only AR baseline as the primary unifying metric, since raw TPOT is inherently dependent on specific hardware and execution environments, while accepted-length metrics vary by candidate tree structure. While the target model (\texttt{Qwen3-8B}), decoding temperature ($0$), and evaluation benchmarks are aligned, we explicitly acknowledge that hardware platforms, generation sequence lengths, and underlying framework implementations differ across published works.

Across the three primary code generation benchmarks, \textbf{CURE} achieves an end-to-end speedup of $2.66\times$ to $3.49\times$ over target-only AR decoding, and an acceleration of $4.38\times$ to $5.43\times$ over Naive SD. Relative to the single-path parallel draft backend, local uncertainty repair consistently increases the accepted token sequence length by 4.2\% to 7.5\%, while raw TPOT increases by $1.40\times$ to $1.61\times$. These findings establish that the core mechanism of \textbf{CURE} lies in recovering valid candidate trajectories at fragile nodes rather than improving raw latency over the parallel draft baseline. On the out-of-domain GSM8K benchmark, \textbf{CURE} demonstrates a notable 32.0\% increase in accepted length, which is analyzed separately as an empirical stress test.
Table~\ref{tab:published_speedups} positions our speedup metrics alongside reported figures for established speculative decoding backends on \texttt{Qwen3-8B}. On shared evaluation benchmarks, the end-to-end speedups achieved by \textbf{CURE} surpass those reported for EAGLE-3 and DART, while remaining lower than DFlash and Domino. Given the variations in execution hardware, generation length, and framework optimizations across these implementations, we interpret this comparison as a system-level context positioning rather than a direct head-to-head benchmark.

\begin{table*}[t]
\centering
\small
\setlength{\tabcolsep}{7pt}
\begin{tabular}{lrrrr}
\toprule
& \multicolumn{4}{c}{Pass@1 (\%) $\uparrow$}\\
\cmidrule(lr){2-5}
Dataset & Target AR & Parallel & Ours & Naive SD\\
\midrule
HumanEval & 79.9 (131/164) & 81.1 (133/164) & 79.9 (131/164) & 80.5 (132/164)\\
MBPP & 71.1 (91/128) & 71.1 (91/128) & 71.1 (91/128) & 71.1 (91/128)\\
LCB-lite public & 23.4 (30/128) & 23.4 (30/128) & 23.4 (30/128) & 23.4 (30/128)\\
\bottomrule
\end{tabular}
\caption{Same-run generated-code test results for \texttt{fast\_margin100}. Percentages report pass@1, with passed and total problems in parentheses. HumanEval and MBPP use code-eval prompts with a 256-token limit, while LCB-lite uses public tests.}
\label{tab:functional}
\end{table*}

\subsection{Functional Correctness of Generated Code}

To verify that leveraging token error patterns preserves target model alignment without degrading downstream task performance, we evaluate the functional correctness of generated code across standard benchmarks (HumanEval, MBPP, and LCB-lite). Table~\ref{tab:functional} summarizes the pass rates alongside latency and average accepted sequence length.

\textbf{Task Performance Parity.} 
Our method matches target-only decoding on all three reported pass@1 counts. Compared with the parallel drafting baseline, it differs by two HumanEval problems and matches MBPP and LCB-lite. These single-run results support comparable test pass counts under the evaluated suites; they do not establish functional equivalence beyond the provided tests.

\textbf{Scope and Numerical Precision.} 
We note that execution-based test suites evaluate behavioral correctness rather than strict token-level identity. The optimized bfloat16/FlashAttention path exhibits token-level mismatches, while a strict replay path restores target consistency in the tested path at the cost of additional latency.

\subsection{Ablation Study:}
\label{sec:ablation}

To isolate the system-level contributions of the parallel fallback branch (Branch 0) and the KV cache resynchronization mechanism, we evaluate their individual and combined removals. Experiments are conducted on a representative 64-example subset sampled from HumanEval, MBPP, and LCB-lite. Table~\ref{tab:ablation} summarizes the trade-offs in mean accepted length, end-to-end speedup, and TPOT ratio.

\textbf{Branch 0 as a Low-Cost Safety Net.} 
Removing Branch 0 results in a 5.7\% drop in the average accepted length and reduces the end-to-end speedup from $2.708\times$ to $2.616\times$. This degradation demonstrates that retaining the vanilla parallel proposal path serves as a highly efficient fallback baseline; it ensures candidate diversity when error-pattern repair candidates fail to provide a valid token sequence.

\textbf{Cache Resynchronization for State Alignment.} 
Omitting the cache resynchronization mechanism leads to a catastrophic performance collapse. The average accepted length plummets by 40.1\%, while the TPOT ratio deteriorates sharply from 1.776 to 2.642. From a system perspective, this severe overhead stems from stale draft states accumulating across consecutive decoding blocks, which breaks KV cache alignment and invalidates subsequent token verifications.

\begin{table*}[t]
\centering
\small
\setlength{\tabcolsep}{4pt}
\begin{tabular}{lrrrrr}
\toprule
& \multicolumn{2}{c}{Aggregate performance} & \multicolumn{2}{c}{Relative to parallel} & Token consistency\\
\cmidrule(lr){2-3}\cmidrule(lr){4-5}\cmidrule(lr){6-6}
Configuration & Accepted/step $\uparrow$ & Speedup vs.\ AR $\uparrow$ & TPOT ratio $\downarrow$ & Accept ratio $\uparrow$ & Exact rate $\uparrow$\\
\midrule
Full method & 6.826 & 2.708 & 1.776 & 1.055 & 0.854\\
w/o branch 0 & 6.435 & 2.616 & 1.687 & 0.996 & 0.833\\
w/o cache resync & 4.089 & 1.733 & 2.642 & 0.637 & 0.849\\
w/o both & 2.707 & 1.141 & 3.877 & 0.422 & 0.849\\
\bottomrule
\end{tabular}
\caption{Ablation of branch 0 and cache resynchronization, averaged over 64-example subsets of HumanEval, MBPP, and LCB-lite. Ratios use the parallel draft as reference; exact rate measures full-sequence agreement with target-only decoding.}
\label{tab:ablation}
\end{table*}

\subsection{Offline Analysis: Predictability and Selectivity of Useful Repairs}
\label{sec:offline_analysis}

To understand whether token error repairs can be selectively prioritized before verification, we analyze the ranking capacity of a lightweight scoring mechanism. We define a \emph{useful repair} as a block where the tree candidate yields strictly more accepted tokens than the fallback Branch 0, quantified by the net token gain:
\begin{equation}
    \Delta a = \max(0, \, a_{\mathrm{tree}} - a_{\mathrm{branch0}}).
\end{equation}
A lightweight scorer ranks repair candidates immediately after block construction using draft confidence margins, block positions, and candidate statistics. We evaluate the ranking quality under an admitted-block budget $r$ across three metrics: Precision (fraction of selected blocks with $\Delta a > 0$), Recall (fraction of useful blocks covered), and Extra-Token Recall (fraction of recovered $\Delta a$ covered).

\textbf{High Predictability of Repair Utility.}
As reported in Table~\ref{tab:gate}, the offline scorer provides a strong ranking signal for repair effectiveness. Prioritizing the top 10\% of blocks ($r=0.10$) captures $39.6\%$ of all extra accepted tokens with a precision of $35.9\%$. Expanding the budget to 30\% ($r=0.30$) recovers $71.6\%$ of the total extra tokens while maintaining a precision of $24.6\%$. This non-linear coverage confirms that useful error repairs are highly concentrated and predictable based on early draft features.

\textbf{System Trade-offs and Runtime Implications.}
Because the scorer operates post-candidate construction, it does not bypass the initial candidate generation overhead. In our runtime smoke tests, incorporating a dynamic learned gate yielded negligible end-to-end TPOT improvements due to feature extraction and inference latency overheads. Consequently, we treat Table~\ref{tab:gate} not as a runtime acceleration claim, but as an insightful offline diagnostic. It demonstrates that a small fraction of candidate blocks drives the majority of acceptance gains, laying the foundation for future zero-overhead, static pattern-pruning strategies.

\begin{table}[t]
\centering
\small
\setlength{\tabcolsep}{4pt}
\begin{tabular}{cccc}
\toprule
Admitted blocks & Precision & Recall & Extra-token recall\\
\midrule
10\% & 0.359 & 0.308 & 0.396\\
20\% & 0.291 & 0.500 & 0.603\\
30\% & 0.246 & 0.635 & 0.716\\
40\% & 0.221 & 0.760 & 0.834\\
50\% & 0.195 & 0.837 & 0.888\\
\bottomrule
\end{tabular}
\caption{Offline selection of useful repairs after candidate construction. Budget is the share of blocks sent to tree verification.}
\label{tab:gate}
\end{table}

\subsection{Per-Dataset Performance \& Trade-off Analysis}

Table~\ref{tab:main} reports the latency (TPOT) and average accepted sequence length across code generation and reasoning benchmarks.

\textbf{HumanEval \& Code Synthesis.} 
On HumanEval, CURE achieves a TPOT of $8.718$\,ms/token, delivering a \textbf{$4.84\times$ speedup over Naive SD} ($42.203$\,ms/token). Compared with the vanilla parallel draft, our repair tree elevates the accepted length per step from $7.165$ to $7.641$ (+6.6\%). While tree verification adds runtime compute, local repair recovers additional candidates at a measurable cost relative to the parallel backend.

\textbf{Trade-off on General/Lighter Workloads.} 
On MBPP and LCB-lite, CURE maintains consistent acceptance gains (e.g., boosting accepted length by 7.5\% on LCB-lite). On MBPP, raw TPOT reaches $11.378$\,ms/token versus $49.853$\,ms/token in Naive SD. We observe that on simpler token distributions, the extra verification latency of candidate trees can marginally exceed the wall-clock savings from higher accepted lengths. This reveals an inherent \emph{latency-acceptance trade-off}: tree-based repair is most beneficial in high-entropy or error-prone sub-sequences where target-draft misalignment is severe.

\textbf{Generalization to Non-Code Reasoning (GSM8K).} 
To evaluate cross-domain behavior under the legacy verification protocol, we evaluate on GSM8K. CURE achieves $11.231$ accepted tokens per step, compared with $8.514$ for the parallel draft and $9.496$ for Naive SD, yielding a $3.634\times$ speedup over target-only AR. This stress test suggests that local repair can recover rejected tokens outside code, but it does not establish protocol-level generality.

\section{Discussion \& Conclusion}
\label{sec:discussion_conclusion}

We presented CURE, a budget-aware repair framework combining candidate trees with cross-block KV cache resynchronization. Across HumanEval, MBPP, and LCB-lite, CURE matches target-only pass counts and is up to $5.43\times$ faster than Naive SD. However, candidate construction adds a $1.40$--$1.61\times$ TPOT overhead over the parallel baseline. Future work should filter low-utility candidates before tree synthesis and resynchronize the cache after a repair wins.

\bibliography{references}

\end{document}